\documentclass{ecai} 

\usepackage{latexsym}
\usepackage{amssymb}
\usepackage{amsmath}
\usepackage{amsthm}
\usepackage{booktabs}
\usepackage{enumitem}
\usepackage{graphicx}
\usepackage{color}

\usepackage{multirow}

\newcommand{\BibTeX}{B\kern-.05em{\sc i\kern-.025em b}\kern-.08em\TeX}

\begin{document}


\begin{frontmatter}


\paperid{0153} 


\title{A Separable Self-attention Inspired by the State Space Model for Computer Vision}


\author[A]{\fnms{Juntao}~\snm{Zhang}\orcid{0000-0001-8174-5378}}
\author[A]{\fnms{Jun}~\snm{Zhou}\thanks{Corresponding Author. Email: JunZhou \_ISE @hotmail.com.}}
\author[B]{\fnms{Kun}~\snm{Bian}} 
\author[B]{\fnms{You}~\snm{Zhou}} 
\author[A]{\fnms{Jianning}~\snm{Liu}} 
\author[C]{\fnms{Pei}~\snm{Zhang}} 
\address[A]{AMS, China}
\address[B]{Xidian University, China}
\address[C]{Coolanyp L.L.C., China}


\begin{abstract}
Mamba is an efficient State Space Model (SSM) with linear computational complexity. Although SSMs are not suitable for handling non-causal data, Vision Mamba (ViM) methods still demonstrate good performance in tasks such as image classification and object detection. We propose a novel separable self-attention method, for the first time introducing some excellent design concepts of Mamba into separable self-attention. To ensure a fair comparison with ViMs, we introduce VMINet, a simple yet powerful prototype architecture, constructed solely by stacking our novel attention modules with the most basic down-sampling layers. Notably, VMINet differs significantly from the conventional Transformer architecture. Our experiments demonstrate that VMINet has achieved competitive results on image classification and high-resolution dense prediction tasks. Code is available at: {https://github.com/yws-wxs/VMINet}.
\end{abstract}

\end{frontmatter}


\section{Introduction}

Modern State Space Models (SSMs) excel at capturing long-range dependencies and reap the benefits of parallel training. The Vision Mamba (ViM) methods \cite{ref1,ref2,ref3,ref4}, which are inspired by recently proposed SSMs \cite{ref5,ref6}, utilize the Selective Space State Model (S6) to compress previously scanned information into hidden states, effectively reducing quadratic complexity to linear. Many studies integrate the original SSM framework from Mamba into their foundational models to balance performance and computational efficiency. However, Mamba is not the first model to achieve global modeling with linear complexity. Linear attention \cite{ref27} replaces the non-linear softmax function with linear normalization and adds a kernel function to both query and key, allowing for the reordering of computation based on the associative property of matrix multiplication, thereby reducing the computational complexity to linear. Separable self-attention \cite{ref7} is also an early work that replaces the computationally expensive operations (e.g., batch-wise matrix multiplication) in Multi-headed Self-Attention (MHA) with element-wise operations (e.g., summation and multiplication). However, because of the limited expressive capabilities of separable self-attention and its variants, they are typically suitable for lightweight vision Transformers that have been carefully designed.\par
Previous studies on ViM have identified a fundamental contradiction between the non-causal characteristics of 2D spatial patterns in images and the causal processing framework of SSMs. Flattening spatial data into 1D tokens destroys the local 2D dependencies in the image, thereby impairing the model's capacity to accurately interpret spatial relationships. Vim \cite{ref1} addresses this issue by scanning in bidirectional horizontal directions, while VMamba \cite{ref2} adds vertical scanning, enabling each element in the feature map to integrate information from other locations in different directions. Subsequent works, such as LocalMamba \cite{ref3} and EfficientVMamba \cite{ref4}, have designed a series of novel scanning strategies. These efforts aim to expand the receptive field of the SSM from the previous token to others, which may result in a multiple-fold increase in the computational cost of the scanning process. Macroscopically, we attribute the success of ViMs to the combination of global information modeling and the establishment of local dependencies, unified by a well-designed architecture. \par
In this paper, we first establish design principles by analyzing the strengths and weaknesses of separable self-attention, classical softmax self-attention, and SSMs. We then confine the receptive field of separable self-attention to the previous token. Furthermore, we introduce the recursive form of our proposed separable self-attention, thereby expressing both SSMs and our method within a unified framework. We refer to this method as the Vision Mamba Inspired Separable Self-Attention (VMI-SA). Finally, we restore the receptive field of our VMI-SA to maintain the advantages of separable self-attention in parallel computing. We construct a demonstrative network, VMINet, by stacking VMI-SA with down-sampling layers. Clearly, the structure of VMINet has not been carefully designed, and it does not adhere to the conventional architectural design principles of the Transformer. For a fair comparison, we keep the number of VMI-SAs consistent with the number of Mamba blocks in Vim \cite{ref1}, and the parameters are roughly equivalent. Experiments demonstrate that our VMINet consistently outperforms Vim and is also competitive with other state-of-the-art models.

\section{Preliminaries}
This section briefly reviews the basic forms of Self-Attention, Separable Self-Attention, and Structured State Space Model.\par
\subsection{Softmax Self-Attention}
In a broad sense, attention refers to a computational process that assigns scores to each pair of positions within a sequence, allowing each element to ``attend" to other elements. The most widely used and significant variant of attention is the softmax self-attention, which can be defined as:
\begin{equation} \label{eq1}
	Y = softmax(QK^T)\cdot V
\end{equation}
where $Q, K, V \in \mathbb{R}^{(L,D)}$ respectively represent $L$ tokens with $D$ dimensions, each generated by a linear transformation from the input $X\in \mathbb{R}^{(L,C)}$. 
The attention scores between each pair of tokens in $Q$ and $K$ are computed using the dot product operation. Subsequently, interactions are normalized using softmax. Finally, the weighted interactions are multiplied by $V$ using the dot product operation to produce the final weighted output. The pairwise comparison mechanism, realized by computing $QK^T$, results in a quadratic growth in the attention's training cost.\par
\subsection{Separable Self-Attention}
The structure of separable self-attention is inspired by Softmax Self-Attention \cite{ref7}. Similar to softmax self-attention, the input $X\in \mathbb{R}^{(L,C)}$ is processed using three branches: $Q\in \mathbb{R}^{(L,1)}, K \in \mathbb{R}^{(L,D)}$ and $V \in \mathbb{R}^{(L,D)}$. Notably,  $Q$  maps each token in $X$ to a scalar, distinguishing it from the other branches. First, context scores are generated through $Softmax(Q)$. Then, based on broadcasting mechanism, the context scores are then element-wise multiplied with $K$ and the resulting vector is summed over the token dimension to obtain the context vector. Finally, the context vector is multiplied by $V$ using broadcasted element-wise multiplication to spread the contextual information and produce the final output. It can be summarized as:
\begin{equation} \label{eq2}
	Y = \sum_{i=1}^{L}\big( softmax(Q)\odot K \big)_i \odot V
\end{equation}
Here, $\odot$ denotes element-wise multiplication. The process follows the broadcasting mechanism throughout.\par
\subsection{Structured State Space Model}
Structured State Space Sequence Model (S4) is a recent sequence model for deep learning, which is widely related to RNNs, CNNs, and classical SSMs. Their inspiration stems from a specific continuous system that, through an implicit latent state $h\in \mathbb{R}^{(D,L)}$, maps a one-dimensional sequence $x\in \mathbb{R}^L$ to another one-dimensional sequence $y\in \mathbb{R}^L$ \cite{ref8}. The mapping process could be denoted as:
\begin{align}\label{eq3}
	\begin{split}
		h_i &= Ah_{i-1} + Bx_i\\
		y_i &= C^\mathrm{T}h_i 
	\end{split}
\end{align}
where $i\in [1,L]$, $A\in\mathbb{R}^{(D,D)}$, $B\in\mathbb{R}^{(D,1)}$ and $C\in\mathbb{R}^{(D ,1)}$. The Selective State Space Model (S6) adopted by Mamba \cite{ref5} is developed based on it. In this paper, we use the term state space model (SSM) to refer to various variants of SSMs, including S4 and S6.

\begin{figure*}[h]
	\centering
	\includegraphics[width=\textwidth]{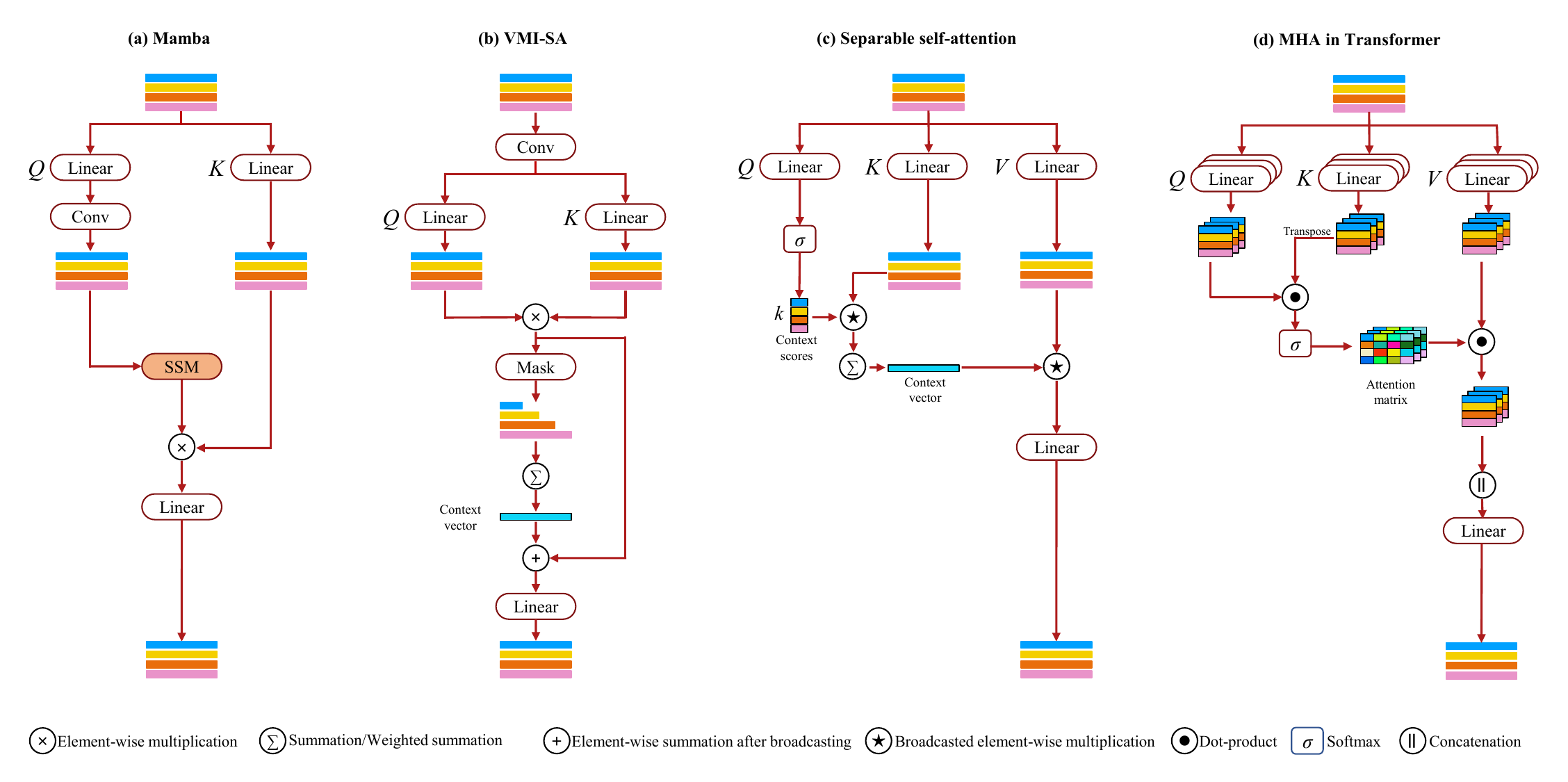}
	\caption{Comparison with different modules. To facilitate a clear comparison, we uniformly adapt one-dimensional sequences as input, although this is not necessary for VMI-SA.}
	\label{Fig1}
\end{figure*}

\section{Methodology}
In this section, we first analyze the impact of the key differences in design between separable self-attention and softmax self-attention. Then, while retaining the advantages of the self-attention design, we optimize the separable self-attention according to the design method of SSM. Our goal is to clearly demonstrate the design process of Vision Mamba Inspired Separable Self-Attention (VMI-SA), to show the innovations and how performance can be enhanced by integrating the strengths of both Mamba and separable self-attention. Finally, we introduce the overall architecture of the proof-of-concept network VMINet.
\subsection{Element-wise Multiplication Instead of Matrix Multiplication}
In both traditional machine learning and deep learning, handling features in high-dimensional space is crucial. We employ a straightforward derivation to establish that both element-wise multiplication and matrix multiplication can map the features from their original dimensions to a higher-dimensional space, which is crucial for feature representation.\par
We adopt the definition method from Section 2, let $X \in \mathbb{R}^{(L,C)}, W^1 \in \mathbb{R}^{(C,D)}, W^2 \in \mathbb{R}^{(C,D)}, Q=X W^1, K=X W^2, E=Q \odot K$. For any element $E_{m,n}$ in $E$ (where $m \in [1,L]$, and $n \in [1,D]$):
\begin{align}\label{eq6}
	\begin{split}
		E_{m,n} &= Q_{m,n} \times K_{m,n} \\
		&= \big(\sum_{i=1}^{C} X_{m,i}  W^1_{i,n} \big) \times \big(\sum_{j=1}^{C} X_{m,j}  W^2_{j,n} \big)\\
		&= \sum_{i=1}^{C}\sum_{j=1}^{C} W^1_{i,n}W^2_{j,n}X_{m,i}X_{m,j}\\
		&= \underbrace {a_{(1,1)}X_{m,1}X_{m,1}+ \cdots + a_{(C,C)}X_{m,C}X_{m,C}}_{C(C+1)/2 \: \mathrm{items}}
	\end{split}
\end{align}
where $a$ is a coefficient for each item:
\begin{align}\label{eq7}
	a_{(i,j)}=
	\left\{
	\begin{aligned}
		&W^1_{i,n}W^2_{j,n} \qquad \qquad \qquad \mathrm{if} \: i==j, \\
		&W^1_{i,n}W^2_{j,n}+W^1_{j,n}W^2_{i,n} \:\:\: \mathrm{if} \: i!=j\\
	\end{aligned}
	\right.
\end{align}
Each term in Eq. (\ref{eq6}) exhibits a nonlinear relationship with the input. It can be approximated as that the element-wise multiplication operation projects the feature vector in the $C$-dimensional space into a higher-dimensional space of $C^2$ dimensions through a nonlinear transformation and processes it.\par
Now let's discuss the case of matrix multiplication. Let $E'=Q \cdot K^{\mathrm{T}}$, where any element $E'_{m,n}$:
\begin{align}\label{eq8}
	\begin{split}
		E'_{m,n} &= \sum_{t=1}^{D} Q_{m,t} \times K^{\mathrm{T}}_{t,n} \\
		&= \sum_{t=1}^{D}\left[\big(\sum_{i=1}^{C} X_{m,i}  W^1_{i,t} \big) \times \big(\sum_{j=1}^{C}  W^2_{j,t}  X_{n,j} \big)\right] \\ 
		&= \sum_{t=1}^{D} \sum_{i=1}^{C}\sum_{j=1}^{C} W^1_{i,t}W^2_{j,t}X_{m,i}X_{n,j} 
	\end{split}
\end{align}
Typically, we consider $D$ to be a constant and $D<<L$. Comparing Eq. (\ref{eq8}) with Eq. (\ref{eq6}), it is evident that from the perspective of information representation, the element-wise multiplication with a linear cost is more efficient than the matrix multiplication with a quadratic cost in terms of computational efficiency.\par
\subsection{Context Vector Instead of Attention Matrix}
The context vector in Eq. (\ref{eq2}) is analogous to the attention matrix $softmax(QK^\mathrm{T})$ in a sense that it also encodes the information from all tokens in the input $X$ \cite{ref7}, but is cheap to compute. Comparing Eq. (\ref{eq6}) and Eq. (\ref{eq8}), it can be observed that $E_{m,n}$ is merely the encoding of the $m$-th token, while $E'_{m,n}$ is the encoding of both the $m$-th and $n$-th tokens. The softmax and summation operations provide a global receptive field for separable self-attention, but the performance difference between separable self-attention and softmax self-attention indicates that establishing correlations between tokens is essential. We speculate that this is also the reason why networks adopting separable self-attention or its variants, such as MobileViT \cite{ref7} and SwiftFormer\cite{ref10}, need to alternately stack the attention modules with local feature encoding modules and feedforward neural network modules. In fact, this perspective is also supported by evidence in ViMs. The SSM restricts the receptive field to the previous token, yet it is still applicable for visual tasks. In addition, it is easy to observe from Eq. (\ref{eq2}) and Eq. (\ref{eq6}) that, due to the parameter sharing across different tokens, the simple summation operation results in identical weights for each token in the global context information, thereby making the computation process of Eq. (\ref{eq2}) lack ``attention". Therefore, in Eq. (\ref{eq2}), the context vector is element-wise multiplied with $V$, which, aside from mapping features to a higher dimension, does not have much clear significance.\par
Additionally, we can analyze the performance differences between softmax self-attention and separable self-attention from the perspective of the rank of the attention matrix. The higher the rank of the attention matrix, the more attention information it contains, and the richer the feature diversity. The attention matrix $softmax(QK^\mathrm{T})$ in Eq. (\ref{eq1}) is usually full rank  \cite{ref36}, that is $\mathrm{rank}(softmax(QK^\mathrm{T}))=L$. The attention information in the context vector comes from $softmax(Q) \odot K$ in Eq. (\ref{eq2}), and its rank:
\begin{align}\label{eq12}
	\begin{split}
		\mathrm{rank}(softmax(Q)\odot K) \leq \mathrm{rank}(K) \leq \mathrm{min}\{ L,D\}.
	\end{split}
\end{align}
Therefore, the attention information in $softmax(Q)\odot K$ is not only less abundant but also severely homogenized.

\subsection{Vision Mamba Inspired Separable Self-Attention}
Summarizing the analysis, the previous discussion provides the following four insights for the design of new separable self-attention:
\begin{itemize}
	\item Continue to use element-wise multiplication for context encoding while reducing the computational branches.
	\item Introduce correlation between tokens.
	\item Enhancing the rank of attention matrices or equivalent counterparts.
	\item Utilize learnable weights to adjust the intensity of each token's contribution to the context information.
\end{itemize}

\subsubsection{Excellent Design in Mamba}
Our analysis results show several similarities with the design philosophies of Mamba. As illustrated in Fig. \ref{Fig1}, for a single Mamba block, the input is processed through two computational branches and then fused via element-wise multiplication, where one branch uses convolution to establish local correlations.\par
In addition, Mamba preserves and compresses global information through the SSM module, which is analogous to the $softmax(QK^\mathrm{T})$ in softmax self-attention mechanism but with linear complexity. As an RNN-based model, Mamba is sensitive to the order of the input sequence, and its scanning process provides the model with positional information. Therefore, unlike transformers, Mamba does not require additional positional encoding. 
\subsubsection{Macro Design}
Our objective is to implement the aforementioned four design philosophies using the simplest and most direct approach, thereby improving the original separable self-attention mechanism without introducing superfluous functional blocks. First, adhering to the design philosophy of separable self-attention, we still utilize context vectors to represent global information. Second, since the contextual information is generated through element-wise multiplication, there is no need to flatten 2D image data into a one-dimensional sequence. Compared to some common Transformers and ViMs, processing features in 2D space can maintain the spatial correlation of features, avoiding the additional inductive bias introduced by Patch Embedding. Additionally, it can reduce the reshaping operations, which is beneficial for improving the inference speed. As previously mentioned, element-wise multiplication can encode the features for individual tokens in pairs, but it cannot establish correlations between tokens. Therefore, the simplest and most effective improvement is to use a depthwise convolution (DW-Conv) layer to establish local spatial correlations before the element-wise multiplication. \par
Next, we consider how to enhance the rank of the attention matrix (or equivalent counterparts). Clearly, for any matrix $A \in \mathbb{R}^{(L,D)}$ with all elements being non-zero, assuming $L>D$, setting the elements of the upper triangular (or lower triangular) part of $A$ to zero can maximize the rank of the matrix, that is:
\begin{equation} \label{eq13}
	\begin{aligned}
		& M = 
		{
			\renewcommand{\arraystretch}{1.5} 
			\setlength{\arraycolsep}{10pt}
			\begin{bmatrix}
				1&     &        &      \\
				1&  1   &       &     \\
				\vdots & \vdots & \ddots &   \\
				1      & 1      & \cdots      &1  \\
				\vdots & \vdots & \vdots & \vdots  \\
				1      & 1      & \cdots      &1  \\
			\end{bmatrix},
		}\\
		&\mathrm{rank}(M \odot A) = \mathrm{min}\{ L,D\} = D, \\
	\end{aligned}
\end{equation}

where $M \in \mathbb{R}^{(L,D)}$. If the matrix $A$ equals the $softmax(Q)\odot K$ from Eq. (\ref{eq2}) and $M$ is regarded as a causal mask matrix, an interesting conclusion can be drawn: the introduction of causality into the separable self-attention can theoretically increase the diversity of contextual information, thereby enhancing performance. Therefore, we believe that it is feasible to improve the separable self-attention by referring to Eq. (\ref{eq3}).\par

\subsubsection{Recurrent Form}
Han et al.\cite{ref9} pointed out that converting linear attention to causal linear attention and introducing a forget gate can significantly improve model performance on ImageNet-1K. It can be observed that in the shallow layers of the network, each token mainly focuses on itself and the two preceding tokens; as the network depth increases, the attention range of each token gradually enlarges. The work of Han et al. indicates that for attention mechanisms with linear computational complexity, the combination of local and global information contributes to forming more effective attention, although their contributions vary at different stages. \par
Like Eq. (\ref{eq3}), we restrict the receptive field to the previous token and preserve past information through a hidden state. The recursive form of the VMI-SA is as follows:
\begin{equation}\label{eq9}
	\begin{aligned}
		h_i &= h_{i-1} + \alpha_i(Q_i \odot K_i)\\
		y_i &= M_i \odot h_i + \beta_i(Q_i \odot K_i)
	\end{aligned}
\end{equation}
where $X \in \mathbb{R}^{(H,W,C)}, W^1 \in \mathbb{R}^{(C,D)}, W^2 \in \mathbb{R}^{(C,D)}, Q=\mathrm{DW-Conv}(X) W^1, K=\mathrm{DW-Conv}(X) W^2$, $L=H*W$, $i\in [1,L]$, $M \in \mathbb{R}^{(L,D)}$ is a lower triangular matrix with all non-zero elements equal to 1, $\alpha_i$ and $\beta_i$  are a series of trainable parameters that control the importance of each token in contextual information, as well as the proportion of local information to contextual information in attention. Like Mamba, we also do not use softmax.
\subsubsection{Matrix Form}
Similar to RNN-based models, the recursive form of VMI-SA is not computationally efficient. The main reason that prevents VMI-SA from being implemented via parallelizable matrix operations is that each token can only utilize information from tokens that precede it in the sequence. Therefore, we remove the restriction on the receptive field and allow all tokens to receive the same global information. Equation (\ref{eq9}) is transformed into:
\begin{equation} \label{eq11}
	\left\{
	\begin{split}
		& Y   = \mathrm{Expand}_L\bigg(\sum_{i=1}^{L} \alpha_i \cdot M_i \odot Q_i \odot K_i\bigg) +\beta \cdot Q \odot K  \\
		& \mathbf{c_v} = \sum_{i=1}^{L} \alpha_i \cdot M_i \odot Q_i \odot K_i
	\end{split}
	\right.
\end{equation}
where $\mathrm{Expand}_L(\cdot)$ denotes the operation of expanding a vector of shape $(1, D)$ into a matrix of shape $(L, D)$, $\mathbf{c_v}$ is the context vector of VMI-SA. The primary network structure of VMI-SA is shown in Fig. \ref{Fig1}. 

\begin{figure}[t]
	\centering
	\includegraphics[width=\columnwidth]{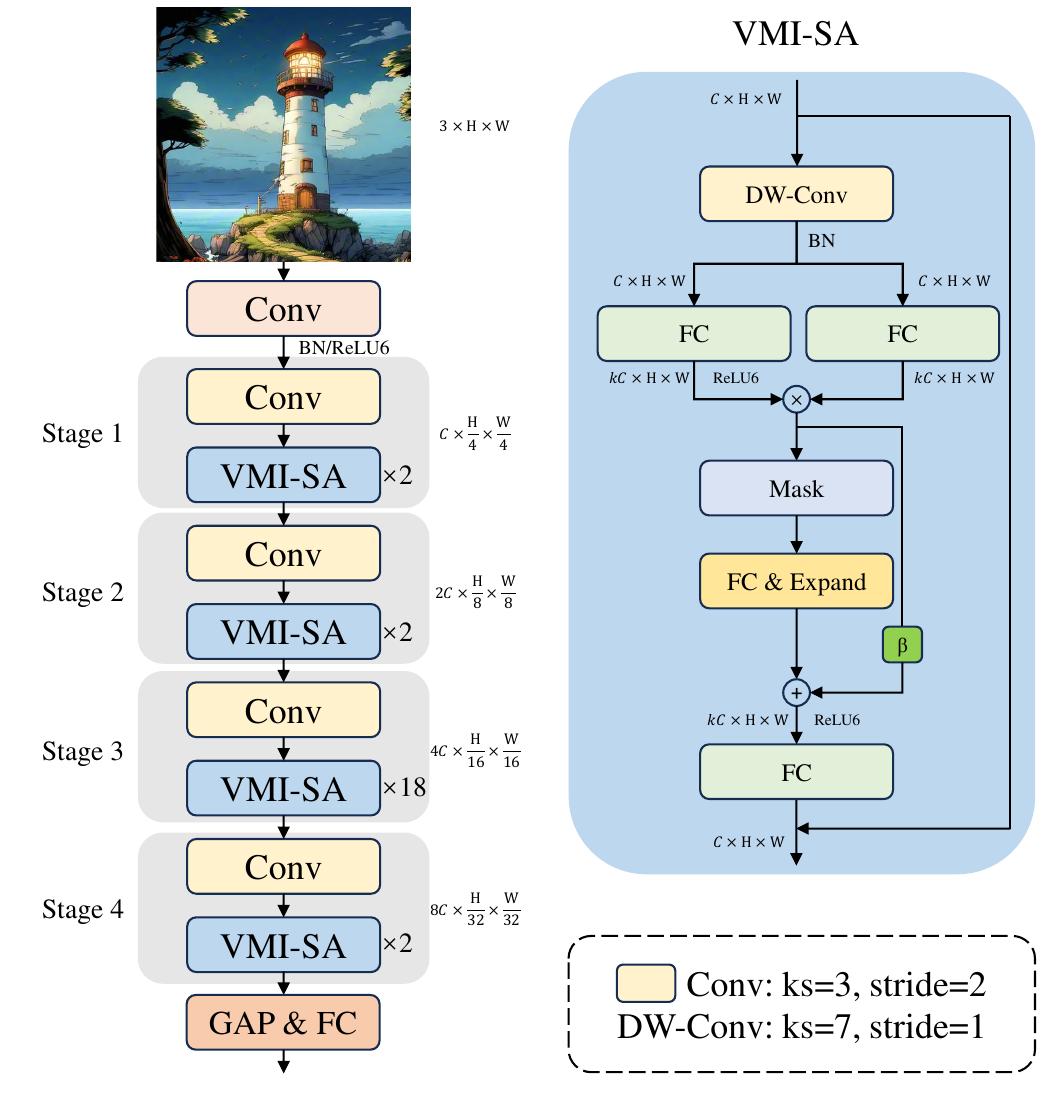}
	\caption{VMINet architecture overview. }
	\label{Fig2}
\end{figure}

\begin{table}[h]
	\centering
	\renewcommand{\arraystretch}{1.5}
	\caption{Configurations of VMINet.}
	\begin{tabular}{cccc}
		\hline
		Variant & \textit{C} & \textit{k} & Params \\
		\hline
		VMINet-Ti & 24 & 2 & 2.0M\\
		VMINet-XS & 48 & 2 & 7.4M\\
		VMINet-S  & 48 & 4 & 13.3M\\
		VMINet-B  & 96 & 2 & 28.4M\\
		\hline
	\end{tabular}
	\label{tab1}	
\end{table}

\subsection{VMINet}
As shown in Fig. \ref{Fig2}, VMINet adopts a common 4-stage hierarchical architecture, utilizing convolutional layers for downsampling, and employing VMI-SA blocks for feature extraction. To ensure a fair comparison with the Vim \cite{ref1}, which uses a pure Mamba encoder, we set the number of VMI-SA blocks to be the same as the number of Mamba blocks with a comparable parameter count. More details can be found in Table \ref{tab1}.

\section{Experiments}
This section presents our experimental results, starting with the ImageNet classification task and then transferring the trained model to various downstream tasks, including object detection, instance segmentation and semantic segmentation.  Additionally, we demonstrate the advantages and disadvantages of VMI-SA variants through comparative experiments.
\begin{table}[h]
	
	\centering
	\renewcommand{\arraystretch}{1.5}
	\caption{Comparison of different models on ImageNet-1K. $\dagger$: In contrast with most of the work presented in the table, MobileViTv2 utilizes a larger resolution of $256 \times 256$, while SwiftFormer employs knowledge distillation.}
	\begin{tabular}{cccc}
		\hline
		\multirow{2}{*}{Method} & Params & FLOPs  & Top-1 \\
		&  (M)   & (G) & (\%) \\
		\hline
		PVTv2-B0 \cite{ref31} & 3 & 0.6 & 70.5\\
		VMINet-Ti \textbf{ (ours)} & 2 &  0.3 & 70.7 \\
		EfficientViT-M2 \cite{ref40} & 4 &  0.2 & 70.8 \\
		LVT \cite{ref34} & 6 & 0.9 & 74.8\\
		Vim-Ti \cite{ref1} & 7 & 1.5 & 76.1\\
		FasterNet \cite{ref12}& 8 & 0.9 & 76.2 \\
		LocalVim-T \cite{ref3} & 8 & 1.5 & 76.5\\
		MobileOne-S2   \cite{ref38}        & 8 & 1.3 & 77.4 \\
		PlainMamba-L1 \cite{ref37} & 7 & 3.0 & 77.9\\
		StarNet-S4 \cite{ref39}     & 8 & 1.1 & 78.4 \\  
		VMINet-XS \textbf{ (ours)}  & 7 & 1.4 & 78.6\\
		EfficientVMamba-S \cite{ref4}  & 11 & 1.3 & 78.7 \\
		DeiT-S \cite{ref32}  & 22 & 4.6 & 79.8\\
		RegNetY-4G \cite{ref20}  & 21 & 4.0 & 80.0\\
		Vim-S \cite{ref1} & 26 & 5.1 & 80.5\\
		VMINet-S \textbf{ (ours)}  & 13 & 2.3 & 80.5\\
		LocalVim-S\cite{ref3}  & 28 & 4.8 & 81.0\\
		Swin-T \cite{ref22} & 29 & 4.5 & 81.3\\
		PlainMamba-L2 \cite{ref37} & 25 & 8.1 & 81.6\\
		ConvNeXt-T \cite{ref41}    & 29 & 4.5 & 82.1\\ 
		VMamba-T\cite{ref2} & 30 & 4.9 & 82.2 \\		
		VMINet-B \textbf{ (ours)}  & 28 & 4.8 & 82.4\\
		\hline
		MobileViTv2-0.5$\dagger$  \cite{ref7} & 1   & 0.5 &  70.2 \\	
		MobileViTv2-1.0$\dagger$  \cite{ref7} & 5   & 1.8 &  78.1 \\	
		SwiftFormer-S$\dagger$  \cite{ref10} & 6 & 1.8   & 78.5\\
		MobileViTv2-1.5$\dagger$  \cite{ref7} & 11   & 4.0 &  80.4 \\		
		SwiftFormer-L1$\dagger$  \cite{ref10} & 12 & 3.2   & 80.9\\
		\hline
	\end{tabular}
	\label{tab2}
\end{table}

\subsection{Image Classification on ImageNet-1K}
\textbf{Settings. }We train the models on ImageNet-1K and evaluate the performance on ImageNet-1K validation set. For fair comparisons, our training settings mainly follow Vim \cite{ref1}. Specifically, we apply random cropping, random horizontal flipping, label-smoothing regularization, mixup, and random erasing as data augmentations. When training on $224\times 224$ input images, we employ AdamW  with a momentum of 0.9 and a weight decay of 0.025 to optimize models. During testing, we apply a center crop on the validation set to crop out $224 \times 224$ images. We train the VMINet models for 300 epochs using a cosine schedule. Unlike Vim, our experiments are performed on 3 A6000 GPUs. Therefore, we adjusted the total batch size and the initial learning rate to 384 and $5 \times 10^{-4} $ respectively.\par
\textbf{Results. }We selected advanced CNNs, ViTs, and ViMs with comparable parameters and computational costs in recent years to compare with our method, and the results are shown in Table \ref{tab2}. The various variants of VMINet are identical in every aspect except for the difference in embedding width. The experimental results demonstrate that VMINet overwhelmingly outperforms Vim \cite{ref1}, which utilizes a pure Mamba encoder. PlainMamba\cite{ref37} has two variants, L1 and L2, which adopt the same configuration of 24 blocks as Vim and VMINet, and employ depthwise convolutions to establish local correlations before selective scanning. Compared with PlainMamba, our VMINet exhibits significant advantages in terms of performance, efficiency, and model complexity. This suggests that VMI-SA is more suitable for visual tasks than Mamba. Furthermore, although VMINet is a demonstrative network architecture that has not been meticulously designed, it still achieves competitive results across various scales, particularly in lightweight scenarios. This suggests that the analysis presented in the previous sections may serve as a guiding principle, potentially reducing the unnecessary attempts researchers might make when designing attention mechanisms or general visual backbone networks.\par
\begin{figure}[h]
	\centering
	\includegraphics[width=\columnwidth]{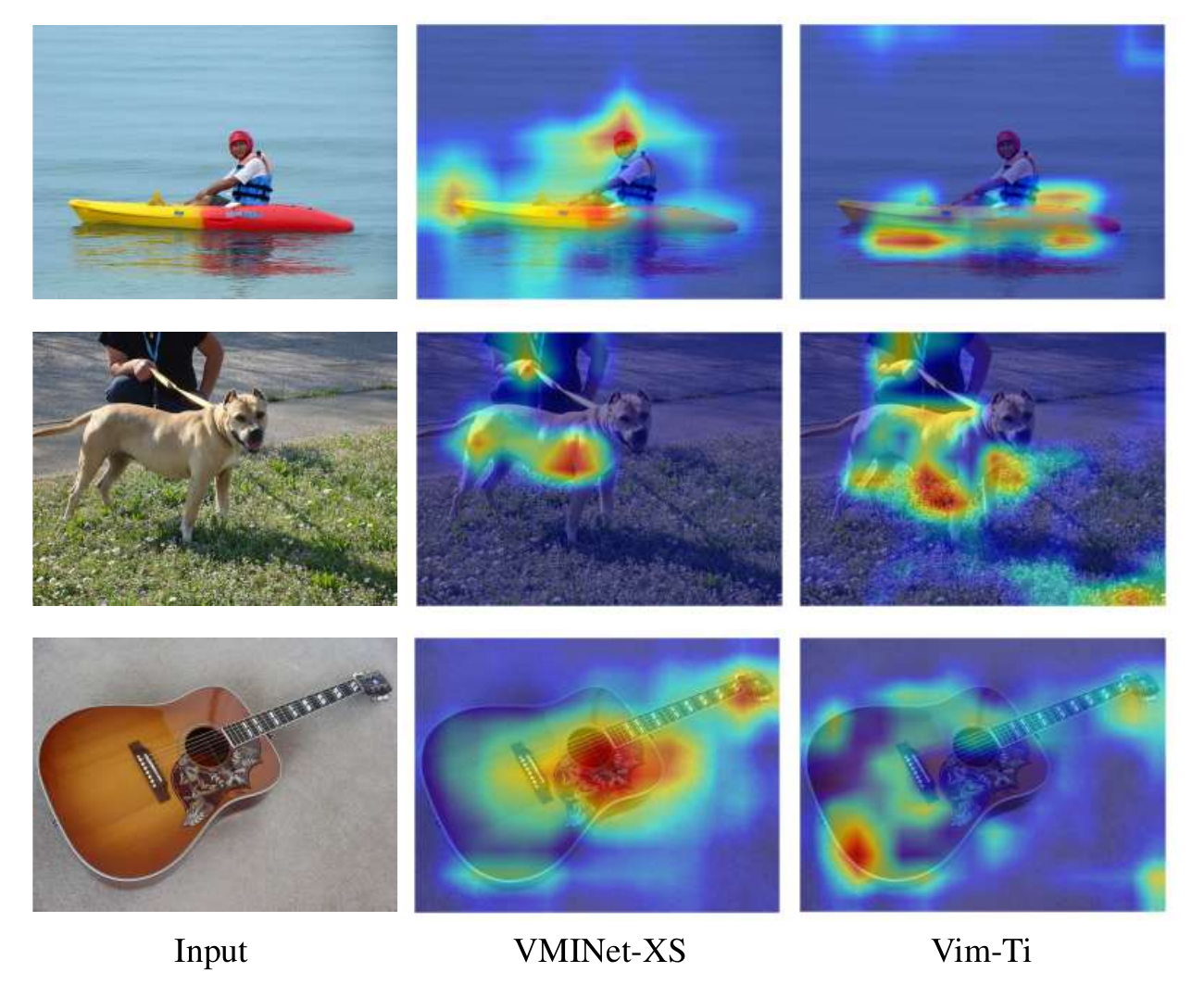}
	\caption{Grad-CAM activation maps of the models trained on ImageNet-1K. The visualized images are from validation set.}
	\label{Fig5}
\end{figure}

We also use Grad-CAM \cite{ref44} to visualize the results of our VMINet-XS and Vim-Ti \cite{ref1} trained on ImageNet-1K. As shown in Fig. \ref{Fig5}, the activation regions of Vim in the maps are more scattered than those of VMINet, and some background areas located at the edges of the image are also activated. Although VMINet also activates some areas outside the classification objects, these regions generally contain certain semantic object information, such as the red helmet.

\subsection{Empirical studies on ImageNet-1K}

\textbf{Recurrent form vs. matrix form.} 
Given that the computational complexity difference between the matrix form and the recurrent form of VMI-SA is negligible, we use latency to measure the actual runtime efficiency difference between them. For comparison, we also report the results of MobileViTv2 \cite{ref7}, SwiftFormer \cite{ref10}, and StarNet \cite{ref39}. Among them, MobileViTv2 and SwiftFormer employ separable self-attention and its variants, while StarNet is a SOTA lightweight model. As shown in Table \ref{tab5}, although VMINet-XS has higher FLOPs than StarNet-S4, the latency of VMINet-XS-M is comparable to that of StarNet-S4. We  believe this is mainly due to the fact that StarNet uses more depthwise convolutions, which significantly increase memory access costs. In terms of performance, VMINet-XS-R slightly outperforms VMINet-XS-M, which can be attributed to the recurrent form of VMINet better utilizing local information across different scales. Considering the trade-off between performance and efficiency, we conclude that the matrix form of VMINet remains a better choice.
\begin{table}[h]
	\centering
	\renewcommand{\arraystretch}{1.5}
	\caption{Comparison of efficient models on ImageNet-1K. The latency is evaluated on an A6000 GPU with a batch size of 1.}
	\begin{tabular}{cccc}
		\hline
		\multirow{2}{*}{Method} & Params & Latency  & Top-1 \\
		&  (M)   & (ms) & (\%) \\
		\hline
		Vim-Ti \cite{ref1} & 7 & 2.6 & 76.1\\
		MobileViTv2-1.0  \cite{ref7} & 5   & 2.3 &  78.1 \\	
		StarNet-S4 \cite{ref39}     & 8 & 1.7 & 78.4 \\  
		SwiftFormer-S  \cite{ref10} & 6 & 2.2   & 78.5\\
		VMINet-XS-M \textbf{ (ours)}  & 7 & 1.8 & 78.6\\
		VMINet-XS-R \textbf{ (ours)}  & 7 & 2.3 & 78.8\\
		\hline
	\end{tabular}
	\label{tab5}
\end{table}
\par

\begin{figure}[h]
	\centering
	\includegraphics[width=\columnwidth]{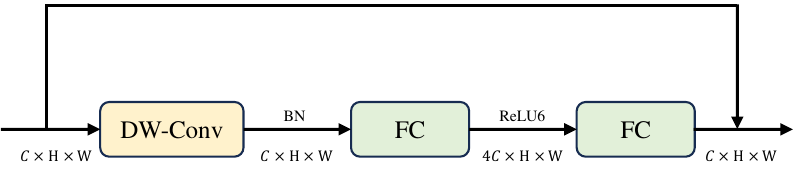}
	\caption{The VMI-SA after removing attention-related operations. It can be observed that it shares the same overall structure as the ConvNeXt block, but differs in normalization methods and activation functions.}
	\label{Fig4}
\end{figure}

\textbf{Effectiveness of VMI-SA.}
Setting aside the design philosophy, due to structural similarities, a reasonable suspicion is that the superior performance of VMINet may primarily be attributed to the introduction of depthwise separable convolutions. As shown in Fig. \ref{Fig4}, for VMINet-S, after removing attention-related operations such as element-wise matrix multiplication and context vector generation, VMI-SA degenerates into a block similar to a ConvNeXt block \cite{ref41}. Although this slightly reduces the number of parameters and computational complexity, the accuracy decreases from 80.5\% to 78.3\%.
\par

\textbf{Impact of mask matrices.}
For matrix-form VMI-SA, the mask matrix $M$ provides positional information for the context vector $\mathbf{c_v}$ while introducing an inductive bias regarding the importance of tokens. Specifically, let $X \in \mathbb{R}^{(L,C)}, W^1 \in \mathbb{R}^{(C,D)}, W^2 \in \mathbb{R}^{(C,D)}, Q=X W^1, K=X W^2, M \in \mathbb{R}^{(L,D)}$. For any element $e_n$ in $\mathbf{c_v}$:
\begin{align}\label{eq14}
	\begin{split}
		e_n &= \sum_{t=1}^{L} \alpha_t \cdot M_t \odot Q_t \odot K_t\\
		&= \sum_{t=n}^{L}\sum_{i=1}^{C}\sum_{j=1}^{C} \alpha_tW^1_{i,n}W^2_{j,n}X_{t,i}X_{t,j}\\
	\end{split}
\end{align}
It is clear that $e_n$ encodes the $n$-th token and all subsequent tokens in the sequence, which implies that tokens with higher indices are encoded more frequently. To balance the importance of each token, the most straightforward method is to remove the mask matrix. However, this leads to a significant performance degradation, with the accuracy dropping from 78.6\% to 76.5\%, primarily due to the loss of positional information. Similar to triangular matrices, banded matrices and block diagonal matrices are also sparse matrices. Using them as mask matrices can provide positional information for $c_v$ while partially alleviating the issue of encoding imbalance. The forms of these matrices are illustrated in Eq. (\ref{eq15}) and Eq. (\ref{eq16}). 
\begin{equation} \label{eq15}
	\begin{aligned}
		&M^1 = 
		{
			\setlength{\arraycolsep}{9.5pt}    
			\begin{bmatrix}
				1 & \cdots & 1 &   &      \\
				\vdots & \ddots  & \vdots  & \ddots  &      \\
				\vdots &   & 1 &   &   1   \\
				\vdots &   & \vdots  & \ddots  &  \vdots   \\
				1 &   & \vdots  &   &  1   \\
				&   \ddots & \vdots  &  & \vdots    \\
				&   & 1  & \cdots & 1  \\
			\end{bmatrix} 
		} 
	\end{aligned}
\end{equation}

\begin{equation} \label{eq16}
	\begin{aligned}
		&M^2 = \begin{bmatrix}
			1      &   \cdots &   1       &        &    &  &\\
			\vdots &   \ddots &   \vdots  &        &  &  &  \\
			1     &  \cdots   &  1  &        &   &  & \\         
			& & & \ddots  & & &   \\
			& & & &   1      &   \cdots &   1        \\
			& & & & \vdots &   \ddots &   \vdots     \\
			& & & &  1     &  \cdots   &  1     \\
		\end{bmatrix} 
	\end{aligned}
\end{equation}

Here, we only discuss two specific cases: Let $M^1,M^2 \in \mathbb{R}^{(L,D)}$, $B=\mathrm{min}(L,D)$, where $M^1$ has a bandwidth of $B/2$, and $M^2$ consists of $B/2$ sub-block matrices, each of size $2 \times 2$. \par
\begin{table}[h]
	\renewcommand{\arraystretch}{1.5}
	\centering
	\caption{Ablation on the impact of different mask matrices.}
	\begin{tabular}{lc}
		\hline
		Form of the Mask Matrix   & Top-1(\%) \\
		\hline
		Baseline   & 76.5 \\
		\hline
		+ Block Diagonal Matrix    & 77.4 \\		
		+ Banded  Matrix           & 78.6\\
		+ Lower Triangular Matrix  & 78.6\\
		+ Hybrid Mask Matrix                 & 78.9 \\
		\hline
	\end{tabular}
	\label{tab6}
\end{table}

\begin{table*}[htbp] 
	\centering  
	\caption{Object detection and instance segmentation results on COCO.} 
	\renewcommand{\arraystretch}{1.4}  
	\begin{tabular}{ccccccccc} 
		\hline
		Backbone & Params & FLOPs & AP$^\mathrm{b}$ & AP${}^\mathrm{b}_{50}$ & AP${}^\mathrm{b}_{75}$ & AP${}^\mathrm{m}$ & AP${}^\mathrm{m}_{50}$ & AP${}^\mathrm{m}_{75}$\\
		\hline
		ResNet-18 \cite{ref15} \ & 31M & 207G & 34.0 & 54.0 & 36.7 & 31.2 & 51.0 & 32.7\\
		Vim-Ti \cite{ref1}     & 27M & 189G & 36.6 & 59.4 & 39.2 & 34.9 & 56.7 & 37.3\\
		PVT-T \cite{ref42} & 33M & 208G & 36.7 & 59.2 & 39.3 & 35.1 & 56.7 & 37.3\\
		ResNet-50 \cite{ref15} & 44M & 260G & 38.0 & 58.8 & 41.4 & 34.7 & 55.7 & 37.2\\
		VMINet-XS   \textbf{(ours)} & 27M & 189G & 38.9 & 61.9 & 42.4 & 36.4 & 58.7 & 38.8\\
		EfficientVMamba-S \cite{ref4} & 31M & 197G & 39.3 & 61.8 & 42.6 & 36.7 & 58.9 & 39.2\\ 
		ResNet-101 \cite{ref15}       & 63M & 336G & 40.0 & 60.5 & 44.0 & 36.1 & 57.5 & 38.6\\
		Vim-S \cite{ref1}            & 44M & 272G & 40.9 & 63.9 & 45.1 & 37.9 & 60.8 & 40.7\\
		Swin-T \cite{ref22}           & 48M & 267G & 42.7 & 65.2 & 46.8 & 39.3 & 62.2 & 42.2\\ 
		VMINet-S   \textbf{(ours)}  & 32M & 201G & 43.2 & 65.3 & 47.3 & 39.3 & 62.2 & 42.3\\
		ConvNeXt-T \cite{ref41}    & 48M & 262G & 44.2 & 66.6 & 48.3 & 40.1 & 63.3 & 42.8\\
		VMamba-T\cite{ref2}        & 48M & 276G & 44.3 & 65.2 & 49.5 & 40.3 & 62.8 & 43.9\\
		VMINet-B   \textbf{(ours)} & 48M & 276G & 44.5 & 66.7 & 48.6 & 40.5 & 63.7 & 43.7\\
		\hline
	\end{tabular}  
	\label{tab4}	
\end{table*}
We use VMINet-XS without the mask matrix as the baseline model and apply different types of mask matrices to it separately. As shown in Table \ref{tab6}, even when using a highly sparse block diagonal matrix as the mask matrix, the model performance still shows a significant improvement. Experimentally, there is no difference in performance when using a banded matrix or a lower triangular matrix as the mask matrix. In addition, we explore the hybrid use of mask matrices. Specifically, the VMI-SA blocks in Stage 1 and Stage 2 use a banded matrix as the mask matrix, while the VMI-SA blocks in Stage 3 and Stage 4 alternately use lower triangular and banded matrices as the mask matrices. The experimental results demonstrate that the hybrid use of different types of mask matrices can achieve better performance. We speculate that carefully designed mask matrices can further enhance the performance of VMINet, and both structural design and parameterization are promising research directions.

\subsection{Object Detection and Instance Segmentation on COCO}
\textbf{Settings. }We use Mask-RCNN as the detector to evaluate the performance of the proposed VMINet for object detection and instance segmentation on the MSCOCO 2017 dataset. Following ViTDet\cite{ref35}, we only used the last feature map from the backbone and generated multi-scale feature maps through a set of convolutions or deconvolutions to adapt to the detector. The remaining settings were consistent with Swin\cite{ref22}. Specifically, we employ the AdamW optimizer and fine-tune the pre-trained classification models (on ImageNet-1K) for both 12 epochs (1$\times$ schedule). The learning rate is initialized at $1\times 10^{-4}$ and is reduced by a factor of $10\times$ at the 9th and 11th epochs.\par
\textbf{Results. }We summarize the comparison results of VMINet with other backbones in Table \ref{tab4}. It can be seen that our VMINet consistently outperforms Vim. Similar to the results on classification tasks, VMINet achieves a good balance between the number of parameters and computational cost, achieving comparable results with advanced CNNs and ViTs.

\subsection{Semantic Segmentation on ADE20K}
\textbf{Settings. }Following Vim \cite{ref1}, we train UperNet \cite{ref43} with our VMINet on ADE20K dataset. In training, we employ AdamW with a weight decay of
0.01, and a total batch size of 16 to optimize models. The employed training schedule uses an initial learning rate of $6 \times 10^{-5}$, linear learning rate decay, a linear warmup of 1500 iterations, and a total training of 160K iterations. \par
\begin{table}[htbp] 
	\renewcommand{\arraystretch}{1.4}  
	\centering  
	\caption{Results of semantic segmentation on ADE20K.} 
	\begin{tabular}{ccc} 
		\hline
		Backbone & Params & mIoU\\
		\hline
		ResNet-50 \cite{ref15}      & 67M & 40.7 \\
		Vim-Ti \cite{ref1}          & 34M & 41.0 \\
		VMINet-XS   \textbf{(ours)} & 34M & 42.7 \\
		Vim-S \cite{ref1}           & 57M & 44.1 \\
		Swin-T \cite{ref22}         & 60M & 44.4 \\ 
		VMINet-S   \textbf{(ours)}  & 47M & 44.8 \\
		ConvNeXt-T \cite{ref41}     & 60M & 46.7 \\
		VMINet-B   \textbf{(ours)}  & 53M & 47.2 \\
		\hline
	\end{tabular}  
	\label{tab7}	
\end{table}
\textbf{Results. }The results are presented in Table \ref{tab7}. Compared with Vim \cite{ref1}, VMINet once again demonstrates higher accuracy and outperforms models such as ResNet \cite{ref15}, Swin\cite{ref22}, and ConvNeXt \cite{ref41}, further validating the effectiveness of VMI-SA.

\section{Conclusion}
This paper presents a separable self-attention inspired by the visual Mamba  (VMI-SA), with linear complexity. Through analysis and derivation, we demonstrate that the element-wise multiplication operation used by separable self-attention can also map the original features to a high-dimensional space for processing, which is more efficient than the matrix multiplication operation used by the classical softmax self-attention. Inspired by the Mamba design philosophy, we first establish local relevance through depthwise convolution, then limit the receptive field to the previous token, and then integrate local and global information according to the recursive state-space model to derive the recurrent form of VMI-SA. Considering the efficiency of matrix operations, we restore the global receptive field and present the matrix form of VMI-SA. We believe that our work can provide a new perspective for the design of future attention mechanisms, that is, by changing the expression and constraints under a unified theoretical framework, to integrate the advantages of different methods. Currently, the research on VMI-SA is still in its infancy, and we believe that with reasonable network structure design, VMI-SA can further improve performance. In addition, the recursive form of VMI-SA is suitable for processing causal data and may be able to compete with other advanced methods in other fields.


\bibliography{mybibfile}

\end{document}